\documentclass[runningheads]{llncs_arxiv}
\usepackage{graphicx}
\usepackage{comment}
\usepackage{amsmath,amssymb} 
\usepackage{color}
\usepackage{multirow}

\begin{document}
\pagestyle{headings}
\mainmatter
\def\ECCVSubNumber{3863}  

\title{Expanding Sparse Guidance for Stereo Matching}

%
\author{Yu-Kai Huang \and
Yueh-Cheng Liu \and
Tsung-Han Wu \and Hung-Ting Su \and Winston H. Hsu}
\authorrunning{Huang et al}
%
%
\institute{National Taiwan University
\email{}}
\maketitle

\begin{abstract}

The performance of image based stereo estimation suffers from lighting variations, repetitive patterns and homogeneous appearance. Moreover, to achieve good performance, stereo supervision requires sufficient densely-labeled data, which are hard to obtain. 
In this work, we leverage small amount of data with very sparse but accurate disparity cues from LiDAR to bridge the gap. We propose a novel sparsity expansion technique to expand the sparse cues concerning RGB images for local feature enhancement. The feature enhancement method can be easily applied to any stereo estimation algorithms with cost volume at the test stage. Extensive experiments on stereo datasets demonstrate the effectiveness and robustness across different backbones on domain adaption and self-supervision scenario. Our sparsity expansion method outperforms previous methods in terms of disparity by more than 2 pixel error on KITTI Stereo 2012 and 3 pixel error on KITTI Stereo 2015. Our approach significantly boosts the existing state-of-the-art stereo algorithms with extremely sparse cues.

\end{abstract}

\section{Introduction}
\begin{figure}[t]
    \centering
    \includegraphics[width=\textwidth]{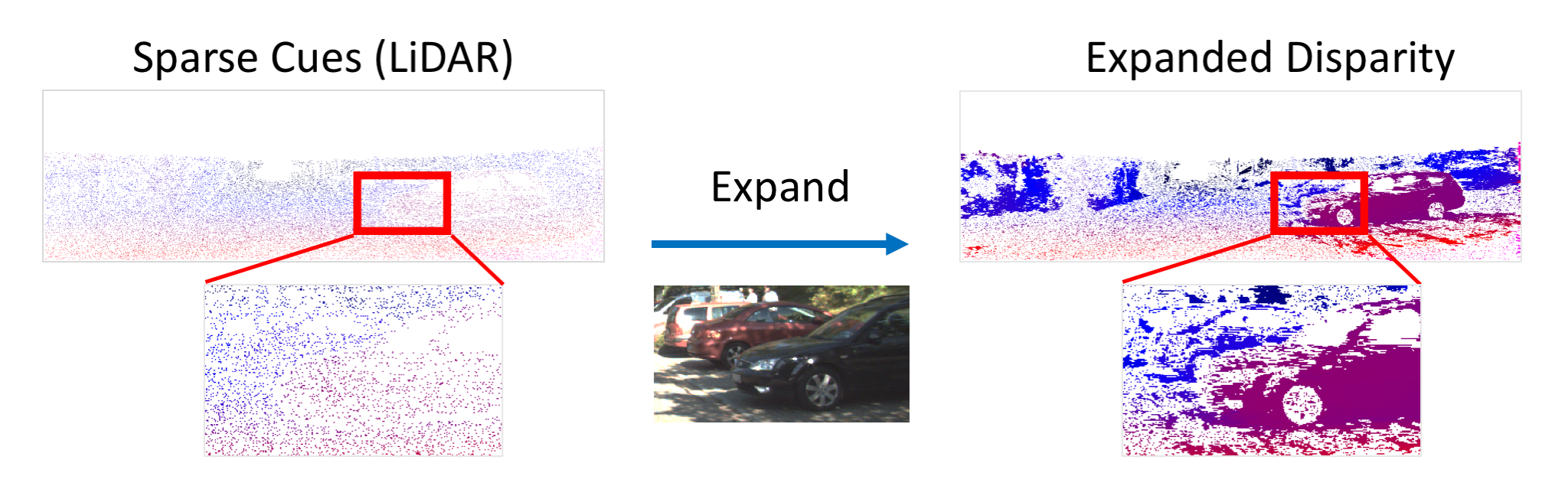}
    \caption{We introduce sparsity expansion, which generates the expanded disparity (right) from the sparse cues (left) with the guidance of RGB images (middle). 
    The expanded disparity maps help improve stereo matching (Section~\ref{subsec:sparsity expansion}).
    }
    \label{fig:fig-1}
\end{figure}

Stereo depth estimation has been widely used and studied for the field of 3D reconstruction \cite{geiger2011stereoscan}, 3D object detection \cite{wang2019pseudo,you2019pseudo} and robotic vision \cite{marapane1989region,nalpantidis2010stereo}. Modern deep learning stereo methods \cite{chang2018pyramid,zhang2019ga} achieve promising performance recently. However, image based matching methods are easily effected by environmental lighting and repetitive textures. In addition, deep learning method via supervised training heavily relies on large-scale dense ground truth data, which is extremely expensive and hard to obtain. In practice, it is impossible to have densely labeled data available. Therefore, it is common to use Scene Flow synthetic data \cite{mayer2016large} for pretraining deep stereo networks.

LiDAR sensor, on the other hand, is well-known for its high accuracy and robustness to weather and lighting; but the sparsity might become an issue. 
As a result, combining the complementary property of stereo dense maps and accurate sparse LiDAR points is an ongoing research trend \cite{maddern2016real,park2018high,wang20193d,poggi2019guided}.

Fusing LiDAR information and RGB image into neural networks for depth estimation is indeed a desirable approach. However, adding new inputs or designing new fusion modules in learning based methods require additional training process. Different from the setting, we aim to directly utilize the sparse LiDAR data as guidance for existing stereo matching algorithms. Poggi et al. \cite{poggi2019guided} proposes feature enhancement method which can be applied on the cost volume of deep stereo networks. However, because the enhancement is confined to sparse pixels, other area can hardly receive gains. The improvement is thus strongly related to the density of external cues.

In this work, we include the effective area of each sparse guidance into consideration and propose the sparsity expansion technique. In this way, the sparse data can be well leveraged to guide stereo matching process. The method can be easily adapted to other stereo algorithms. With two different deep stereo backbones, we experiment our method on three different settings, including domain transfer from synthetic data to real stereo dataset, domain transfer with additional fine-tuning, and self-supervised task. The result demonstrates significant improvements, outperforms previous methods and further proves the effectiveness of sparsity expansion and weighted distance. More precisely, the domain transfer and self-supervised experiments prove that without ground truth data during training, we can boost the performance and reduce the error gap compared to fully supervision with sparse external cues available, (e.g., low-cost LiDAR). Our main contributions are highlighted as follows,
\begin{itemize}
  \item We formulate the sparse guided stereo matching problem that took the effective areas of the sparse guidance into consideration.
  \item We propose the sparsity expansion and distance weighted feature enhancement that propagates the guidance into neighbor areas .
  \item Our method shows generalizability over multiple backbone networks, datasets, and experiment settings.
\end{itemize}

\section{Related Work}
Estimating depth by stereo image pair has been a widely studied research topic. \cite{scharstein2002taxonomy} divides the stereo matching problem into four steps: cost computation, cost aggregation, disparity optimization/computation, and disparity refinement. \cite{hosni2012fast} propose using 3D cost volume that fasten the cost computation and aggregation. On the other hand, \cite{hirschmuller2005accurate} semi-global matching has been largely used due to its trade-off between the speed of local methods and the accuracy of global methods.

Deep learning methods also plays an important role in this field. \cite{vzbontar2016stereo,zbontar2015computing,luo2016efficient,zagoruyko2015learning} compute matching cost with a learned neural network which is trained to predict the similarity of two image patches via Siamese-like loss functions. More recently, end-to-end fashion is adopted to build a powerful stereo matching algorithm that directly output the dense disparity map \cite{mayer2016large,shaked2017improved,kendall2017end,liang2018learning,chang2018pyramid,zhang2019ga}. These end-to-end methods require huge amount of supervised training data. Therefore, synthetic environments \cite{mayer2016large} are often used for pretraining.

In order to achieve better results in the challenging outdoor scenario, there are works combine the information of LiDAR and RGB images for stereo depth estimation \cite{park2018high,wang20193d}. \cite{park2018high} first recovers a dense disparity map using SGM \cite{hirschmuller2005accurate}, which is then fused with LiDAR data and color image in a two-stage fully convolutional network for refinement. \cite{wang20193d} fuses LiDAR data with stereo images at the input phase and adopt conditional batch normalization at the cost volume regularization step which takes LiDAR as condition. 

Besides fusion based methods treat LiDAR data as additional input, LiDAR can be used to directly guide or refine the output. \cite{you2019pseudo} proposes a graph-based depth correction algorithm that refines the pseudo point cloud generated by stereo methods. Yet, the optimization-based correction algorithm of the second stage makes the estimation far away from real-time usage.  \cite{poggi2019guided} applies enhancement on the cost volume features with the depth knowledge from the LiDAR point cloud via Gaussian function. However, only the points with LiDAR data available are enhanced. Different from previous methods, our main idea is to completely exploit the guidance of external sparse but accurate depth information acquired from cheap sensors in the task of stereo disparity estimation.

\section{Method}

\begin{figure}[t]
    \centering
    \includegraphics[width=0.9\textwidth]{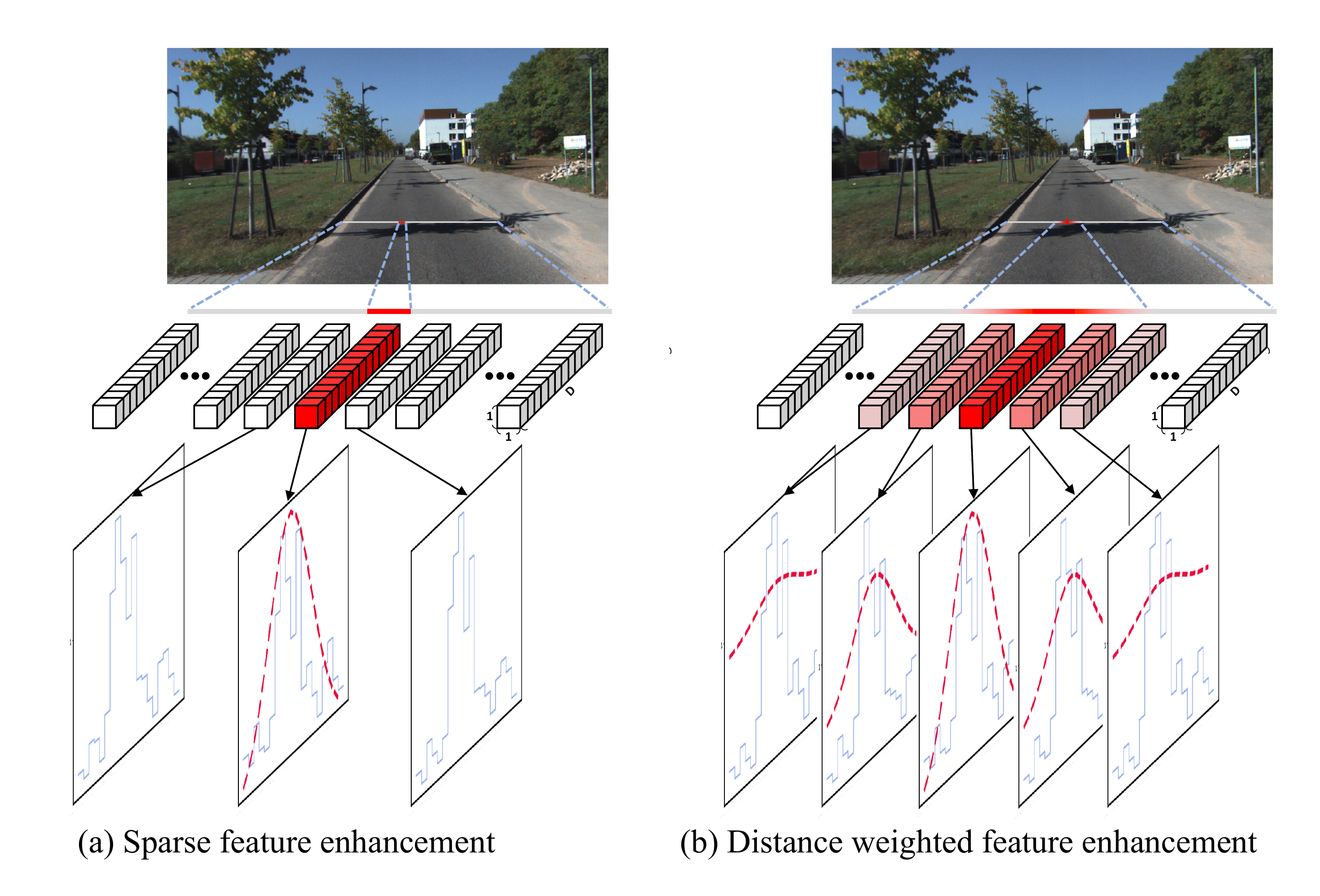}
    \caption{An illustration of distance weighted feature enhancement. We show the  slice of the cost volume with size $H \times W \times D$ along a horizontal line in $D$ dimension vectors. It can be further plotted into figures in blue line. Given a guidance point (red point in image). (a) Poggi et al. \cite{poggi2019guided} apply Gaussian function (red dash line) on the guided location to enhance the features with the guidance point's disparity value. Other pixels without external cues remain the same. (b) Our method propagates the sparse guidance to the neighboring pixels by enhancing the cost volume feature with a distance weighted function that considers the distance to the guidance point. Therefore, pixels without external cues can also be guided by the sparse cues.}
    \label{fig:fig-2}
\end{figure}
In this section, we will first briefly review stereo matching algorithm and the feature enhancement method on the cost volume \cite{poggi2019guided}. Then, we will introduce (1) sparsity expansion and (2) distance weighted feature enhancement. 

In the following, we denote stereo image pair as $I_l, I_r$ with dimensions $H \times W \times 3$. Let $S \in \mathbb{R}^{N \times (2+1)}$ be an external sparse but accurate data, where $N \ll H \times W$, $(2+1)$ dimension denotes the pixel coordinate $x, y$, and the disparity value $d$ for each sparse point. The goal is to compute the dense disparity map with stereo pair $I_l, I_r$ guided by $S_i$ despite its sparsity.

\subsection{Stereo Matching}
Traditional stereo matching algorithms take the stereo image pairs and construct a 3D cost volume \cite{hosni2012fast} to represent the matching cost for each pixel pairs with respect to different disparities. Filtering is then applied along height and width dimension for cost aggregation and smoothing. In recent deep learning stereo matching methods \cite{chang2018pyramid,zhang2019ga,kendall2017end}, following this framework, form cost volumes $C \in \mathbb{R}^{H \times W \times D \times F}$ where $D$ is the maximum disparity value, and $F$ is the feature dimension. 

\subsection{Feature Enhancement}
The naive way to leverage the sparse guidance would be directly replacing the output disparity value at that point with the more confident one. For the cost volume, this can be done by setting the cost of the specific disparity value in cost volume to zero. However, the hard assignment and zeros may damage the deep learning model. Instead, \cite{poggi2019guided} suggests to enhance cost volume feature by multiplying a Gaussian function.

Given the pixel coordinate $x_i, y_i$ and disparity value $d_i$ from external cue $S$, the enhancing function $g$ for the cost volume is
\begin{equation}
    g_i(d) = h \cdot e^{-\frac{(d - d_{i})^2}{2w^2}}
    \label{eq:fe}
\end{equation}
where $h$ and $w$ control the height and width of the Gaussian. This function is directly multiplied with the each entry $(x_i, y_i, d_i) \in S_i$ in the cost volume. The multiplication repeats along the feature dimension, so we will ignore the feature dimension in our discussion.

The method is capable of attaching into most of the stereo methods with cost volume easily. However, the enhancement is only applied to those pixels with sparse cue. This is a considerable issue when the external data is extremely sparse. 

\subsection{Sparsity Expansion}\label{subsec:sparsity expansion}
\begin{figure}
    \centering
    \includegraphics[width=0.7\textwidth]{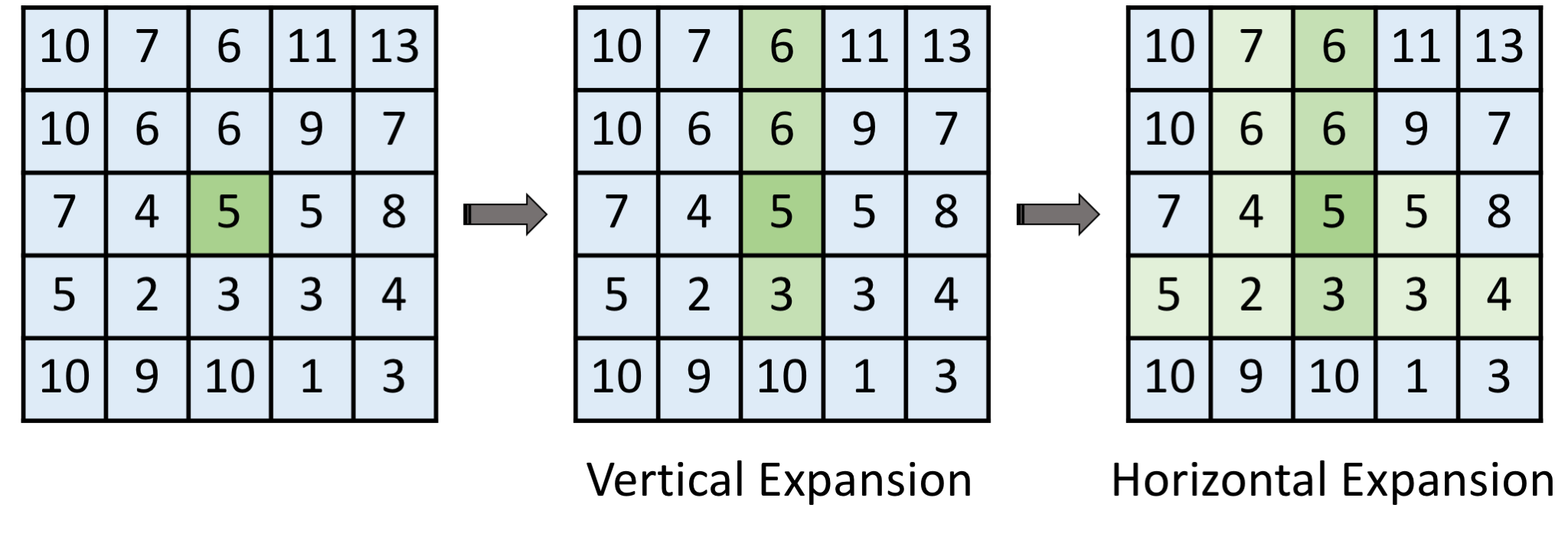}
    \caption{
    An example of expanding the effective area with intensity difference threshold $\tau=3$ and length limit $L=2$. Each number denotes as the intensity value of the pixel. The center point represents the disparity hint from the sparse set. We first expand vertically. The bottom direction stop at $3$ since $|5 - 10| > \tau$. We then search horizontally from each point in the vertical line.}
    \label{fig:fig-3}
\end{figure}
Instead of applying feature enhancement only on the sparse points for guidance, we propose to expand the sparse guidance to the neighbor area. 
Our idea is that the neighbor pixels with same color intensity in the RGB image comes from the same image structure. 
These pixels are all under the effective region of the guidance point even though they do not have explicit guidance. By taking this into consideration, we can expand the sparse guidance to other place and much larger area can be benefited from the external cues.

For a guiding point $S_i = (x_i, y_i, d_i) \in S$ where $x_i, y_i$ are the index of pixel in image and $d_i$ is the disparity value, we aim to find an effective area $G_i$ for such point $S_i$. Specifically, the pixels in $G_i$ are the neighbor pixels of $S_i$ with similar RGB intensity. In contrast, $G_i = \{(x_i, y_i)\}$ in \cite{poggi2019guided}.

Thankfully, there are many existing algorithms that can help us to find $G_i$ for each $(x_i, y_i)$. For example, super-pixel for oversegmentation via clustering based method.
Here, we adopt a relative simple method introduced by \cite{zhang2009cross}. It was originally proposed to generate adaptive support region for local stereo matching. Given a central pixel $(x_i, y_i)$, we greedily search vertically along up and down until the intensity difference between the current location and the center is larger than threshold $\tau$ or the vertical length reaches the limit $L$. For each pixel found in vertical search, search horizontally following the same rule. All the pixels found in the cross shape search are put into $G_i$. Figure~\ref{fig:fig-3} shows an example of expanding the point into an area. The algorithm is able to find a descent effective area that preserves edges via intensity difference thresholding. Moreover, it can be easily parallelized since the expansion process for each point is independent.

\subsection{Distance Weighted Feature Enhancement} \label{sec:distance-weighted}
Naively, we can apply the same $g$ for all the pixels in $G_i$, assuming that they all have the similar depth values. However, the farther distance between $(x, y) \in G_i$ and $(x_i, y_i)$, the less confidence we are to set $(x, y)$ to $d_i$. Not all the pixels in $G_i$ should apply the same level of enhancement as $(x_i, y_i)$. Therefore, we design a new weighting function $f$ that takes distance into consideration. For $(x, y) \in G_i$, 
\begin{equation}
f_i(x, y, d) = \big(1 - \alpha_i(x, y)\big) \cdot g_i(d) +  \alpha_i(x, y) \cdot 1
\end{equation}
where
\begin{equation}
     \alpha_i(x, y) = \min \Big( 1, \frac{|| (x, y) - (x_i, y_i)||}{v} \Big)
\end{equation}
and $v$ is the parameter which determines the sensitivity of the distance. $f$ is a linear interpolation between $g$ and $1$. When $(x, y)$ is very far from $(x_i, y_i)$, $f_i(x, y, d)$ equals to $1$ for all $d \in [0, D]$, which means to keep the original cost volume and not to apply enhancement.
\begin{figure}
    \centering
    \includegraphics[width=\textwidth]{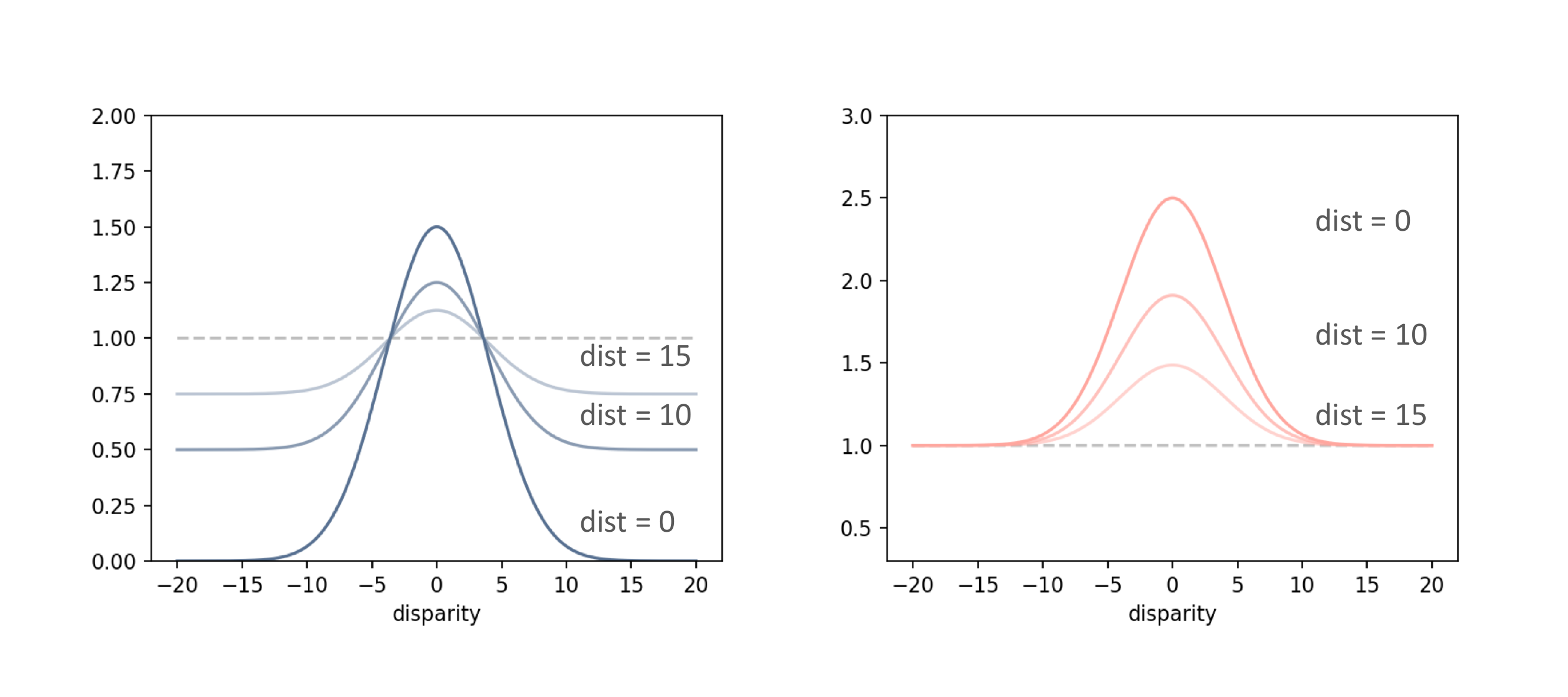}
    \caption{An example plot of the distance weighted function. $f$ (left) and the shifted version $f^s$ (right) given different distance values. See details in Section~\ref{sec:distance-weighted}}
    \label{fig:figure-4}
\end{figure}

We are concerned about the dampening effect of $g$ zeros the features in the cost volume with disparity value different from the disparity hint. It may prevent the gradient from back-propagating for those entries. Hence, we also propose another version of distance weighted function $f^s$,
\begin{equation}
    f^s_i(x ,y, d) = b + h e^{- \left(
    \frac{\left( d - d_i \right)^{2}}{2 w^{2}} 
    + \frac{|| (x, y) - (x_i, y_i)|| ^{2}}{2 v^{2}}\right)}
    \label{eq:fe}
\end{equation}
where $s$ stands for shifted and $b$ can adjust the base height of this function. When $b=1$, $f^s$ is very similar to $f$ except for that it only promote the disparity values close to $S_i$ and does not lower any value. In our experience, $b$ can also be a small number $ > 0$ that lowers other values far from the target disparity value.
Figure~\ref{fig:figure-4} shows us an example of t $f$ and $f^s$ when $|| (x, y) - (x_i, y_i)||$ is equal to $0$, $10$, or $15$. As the distance becomes larger, $f$ and $f^s$ have smaller peaks, which means the enhancement level is reduced.

\section{Experimental Results}
In the section, we perform a series of experiments to validate our feature enhancement methods. To evaluate the improvement on domain adaption, we qualitatively conduct experiments on three datasets and two backbone models and demonstrate similar improvements on self-supervised pretrained model. In addition, we investigate the robustness of our methods concerning the density of sparse cues.

\subsection{Experiment Details}
We select two different yet representative deep stereo networks, PSMNet \cite{chang2018pyramid} and GANet \cite{zhang2019ga}, which source codes are officially available. We use the weights pretrained with SceneFlow \cite{mayer2016large} for $10$ epochs offered by \cite{chang2018pyramid,zhang2019ga} in the official repositories as initialization. 

For our hyper-parameter settings, we set the expansion threshold $\tau = 15$ for feature enhancement, while $\tau = 5$ for fine-tuning. We set the length limit $L$ to $30$ pixels. Note that the sparsity expansion can be performed in highly parallelized and offline, since each pixel is independent. We use grid search to find the best hyper parameters for Gaussian height $h$, width $w$, sensitivity of the distance $v$ and base height $b$. We explicitly denote the value of each hyper-parameters used in experiments in Table~\ref{tab:hyp-param}. For the comparison baseline \cite{poggi2019guided}, we use simple grid search and find consistent results of hyper-parameter values mentioned in his paper $w=1, h=10$, which we use in the following experiments.
\begin{table}
    \centering
    \caption{The hyper-parameters we use to conduct the following experiments. The value of hyper-parameters are found by grid search method.}
    \begin{tabular}{|c|c|c|c|c|c|c|} \hline
         Experiment Name&Fine Tune&Weighted Distance&$w$&$h$&$v$&$b$\\ \hline
         PSMNet, GANet&&$f$&1&20&30&- \\ 
         PSMNet&&$f^s$&1&20&1&0.1\\ 
         GANet&&$f^s$&1&100&10&1\\ \hline
         PSMNet, GANet&\multirow{2}{*}{\checkmark}&$f$&10&1&42&- \\
         PSMNet, GANet&&$f^s$&10&10&15&1\\ \hline
    \end{tabular}
    \label{tab:hyp-param}
\end{table}
\subsubsection{Domain Adaption Setting}
KITTI Stereo 2012 dataset contains 194 training image pairs for left-view image, right-view image and semi-dense disparity maps, which consists of accumulation point clouds from the nearby 10 frames. To obtain sparse disparity maps, we sub-sample 15\% of semi-dense disparity maps, resulting in an average 4.2\% coverage of sparse cues.  For KITTI Stereo 2015, it contains 200 image pairs, and we follow similar procedure in KITTI Stereo 2012, and result in an average coverage of 2.9\%. For both dataset, following the protocol in PSMNet \cite{chang2018pyramid}, we divide it into 80\% and 20\% as training set and validation set for fine-tuning purpose.

\subsubsection{Self-supervision Setting}
We use Eigen \cite{eigen2015predicting} KITTI split, which contains 22600 pairs of training data, 3030 pairs of validation data and 652 pairs of test data. We train the network without using the ground truth data in the training set. When fine-tuning, we divide test set into 80\% and 20\% for fine-tuning and validation. For the self-supervision model, we follow the training protocols and losses in \cite{zhong2017self} and the hyper-parameters from the paper. The only difference is that we reimplement the stereo network with PSMNet \cite{chang2018pyramid} as backbone model due to no public source codes available. 

\subsubsection{Evaluation Metrics} \label{subsubsec:evaluation metrics}
We evaluate on full resolution of all datasets and report the average pixel error of disparity 
$$\text{average pixel error} = \frac{\sum{\vert D_{\text{pred}} - D_{\text{gt}} \vert }}{N}$$
, where $N$ denotes the number of pixels with valid ground truth disparity. For $n$-pixel error, it denotes the percentage of pixels with wrong estimated disparity, where $n \in \{2, 3, 4, 5\}$. Given that a pixel is considered to be correctly estimated if the disparity error $<n$, $n$-pixel error calculates the percentage of pixels predicted wrongly. 

\subsection{Evaluation on KITTI Stereo Datasets}
We evaluate on both KITTI Stereo 2012 and 2015 dataset. The result is shown in Table~\ref{table:kitti-2012} and Table~\ref{table:kitti-2015}. Directly inference on KITTI using the model trained with SceneFlow 
without fine-tuning performs badly for PSMNet. In this case, our method can largely improve the average pixel error from 8.156 to 5.273 in KITTI 2012 and from 8.568 to 4.624 in KITTI 2015. The scores using GANet also show a performance boost.

As for the fine-tuning experiment, "Feature Enhance" means to apply feature enhancement during training and test. Both networks with small set of training data for fine-tuning already can achieve very promising results. Yet, our method still has brought the models an amount of improvement. 

\setlength{\tabcolsep}{1.4mm}{
\begin{table}[t]
    \caption{Evaluation on KITTI stereo 2012 dataset. We can see that our method outperforms guided stereo matching (GSM) \cite{poggi2019guided} in both with and without fine-tuning in both backbones. Backbones are PSMNet and GANet. Fine tune represents fine-tune on the train split and test on validation split. Feature enhancement is to apply the guidance in stereo matching. The detail of evaluation metrics are included in Section~\ref{subsubsec:evaluation metrics}.}
    \centering
    \begin{tabular}{|c|c|c|c||c|c|c|c|c|} \hline
        \multirow{2}{*}{Backbone} & \multirow{2}{*}{Model} & \multirow{2}{*}{\shortstack{Fine\\Tune}} & \multirow{2}{*}{\shortstack{Feature\\Enhance}} & \multirow{2}{*}{Avg PX} & \multicolumn{4}{c|}{Error Rate}\\ \cline{6-9}
        &  &  &  & & $>2$ & $>3$ & $>4$ & $>5$ \\\hline \hline
        \multirow{6}{*}{\shortstack{PSM\\Net\cite{chang2018pyramid}}} & Baseline &  &  &8.156 & 0.788 &  0.680 & 0.577& 0.482 \\ 
        & GSM \cite{poggi2019guided} & & \checkmark & 7.319&0.784&0.678&0.577&0.485 \\ 
        & Ours & & \checkmark & \textbf{5.273}&\textbf{0.753}&\textbf{0.645}&\textbf{0.544}&\textbf{0.449} \\ \cline{2-9} 
        & Baseline & \checkmark & & 0.604 &0.039&0.024&0.018&0.014 \\ 
        & GSM \cite{poggi2019guided} & \checkmark & \checkmark & 0.473 & 0.025 & 0.015 & 0.011 & 0.009 \\ 
        & Ours & \checkmark & \checkmark & \textbf{0.389} & \textbf{0.020} & \textbf{0.011} & \textbf{0.008} & \textbf{0.006} \\ \hline \hline
        \multirow{6}{*}{\shortstack{GA\\Net\cite{zhang2019ga}}} & Baseline & & & 2.881 & 0.432 & 0.236 & 0.157 &0.117 \\ 
        & GSM \cite{poggi2019guided} & & \checkmark & 	2.559&0.240&0.128&0.096&0.079 \\  
        & Ours & & \checkmark & \textbf{1.812}&\textbf{0.232}&\textbf{0.091}&\textbf{0.051}&\textbf{0.032} \\ \cline{2-9} 
        & Baseline & \checkmark & &  0.582 & 0.038 & 0.023 & 0.017 & 0.014  \\  
        & GSM \cite{poggi2019guided} & \checkmark & \checkmark & 0.460 & 0.025 & 0.016 & 0.012 & 0.010 \\ 
        & Ours & \checkmark & \checkmark & \textbf{0.360}  & \textbf{0.015} & \textbf{0.009} & \textbf{0.007} & \textbf{0.006} \\ \hline
    \end{tabular}
    
    \label{table:kitti-2012}
\end{table}
}

\setlength{\tabcolsep}{1.5mm}{
\begin{table}
    \centering
    \caption{Results on KITTI Stereo 2015 dataset. This shows the identical result in KITTI 2012. Our method can reduce the performance drop due to domain shift and boost the fine-tuning result for both networks.}
        \begin{tabular}{|c|c|c|c||c|c|c|c|c|} \hline
        \multirow{2}{*}{Backbone} & \multirow{2}{*}{Model} & \multirow{2}{*}{\shortstack{Fine\\Tune}} & \multirow{2}{*}{\shortstack{Feature\\Enhance}} & \multirow{2}{*}{Avg PX} & \multicolumn{4}{c|}{Error Rate}\\ \cline{6-9}
        &  &  &  & & $>2$ & $>3$ & $>4$ & $>5$ \\\hline \hline
        \multirow{6}{*}{\shortstack{PSM\\Net\cite{chang2018pyramid}}} & Baseline &  &  &8.568 & 0.730 & 0.600 & 0.486 & 0.390 \\ 
        & GSM \cite{poggi2019guided} & & \checkmark & 7.544 & 0.724 & 0.594 & 0.483 & 0.388 \\ 
        & Ours & & \checkmark & \textbf{4.624} & \textbf{0.713} & \textbf{0.584} & \textbf{0.472} & \textbf{0.373} \\ \cline{2-9} 
        & Baseline & \checkmark & & 0.764 & 0.055 & 0.029 & 0.019 & 0.014 \\ 
        & GSM \cite{poggi2019guided} & \checkmark & \checkmark & 0.603 & 0.031 & 0.016 & 0.011 & 0.009 \\ 
        & Ours & \checkmark & \checkmark & \textbf{0.502} & \textbf{0.023} & \textbf{0.012} & \textbf{0.008} & \textbf{0.006} \\ \hline \hline
        \multirow{6}{*}{\shortstack{GA\\Net\cite{zhang2019ga}}} & Baseline & & & 3.238 & 0.494 & 0.277 & 0.175 & 0.124  \\ 
        & GSM \cite{poggi2019guided} & & \checkmark & 2.956 & 0.419 & 0.220 & 0.141 & 0.102 \\  
        & Ours & & \checkmark & \textbf{2.030} & \textbf{0.261} & \textbf{0.091} & \textbf{0.048} & \textbf{0.032}\\ \cline{2-9} 
        & Baseline & \checkmark & &  0.743 & 0.052 & 0.028 & 0.019 & 0.014  \\  
        & GSM \cite{poggi2019guided} & \checkmark & \checkmark & 0.586 & 0.030 & 0.017 & 0.012 & 0.009 \\ 
        & Ours & \checkmark & \checkmark & \textbf{0.506} & \textbf{0.021} & \textbf{0.012} & \textbf{0.009} & \textbf{0.007}\\ \hline
    \end{tabular}
    \label{table:kitti-2015}
\end{table}
}


\subsection{Evaluation with Self-supervised Stereo Algorithm}
This experiment shows that although we can train models in self-supervised fashion without ground truth data, we can further improve the result with additional sparse input ground truth by our method.

Here, we adapt the method from \cite{zhong2017self} which introduce the image warping loss on stereo pair. \cite{zhong2017self} takes the disparity map generated from stereo matching algorithm and warps the left image to synthesised right image to form the image difference loss. We re-implement the method by using PSMNet \cite{chang2018pyramid} as stereo matching backbone and train with image warping loss without ground truth data involved. We then apply our sparsity expansion module during inference time.

The results in Table~\ref{table:self} shows the improvement of self-supervised stereo matching with or without feature enhancement and sparsity expansion. The results are consistent with KITTI Stereo 2012 and KITTI Stereo 2015, and further proves the generalizability of our methods to existing stereo matching tasks.  
\begin{table}
    \centering
    \caption{Evaluation on the self-supervised stereo model. Given sparse cues in testing phase, our feature enhancement method is able to increase the overall performance of the model trained with self-supervised loss. * we re-implement \cite{zhong2017self}.}
    \begin{tabular}{l|c|c|c|c}
        \hline
        Model & Avg PX & $>3$ Error & $>4$ Error &  $>5$ Error\\ \hline
        Self-supervised stereo \cite{zhong2017self}* &  3.136 & 0.213 & 0.171 & 0.142  \\
        w/ GSM \cite{poggi2019guided}&3.108 & 0.197 & 0.158 & 0.134\\
        \textbf{w/ Ours} & \textbf{3.028} & \textbf{0.162} & \textbf{0.126} & \textbf{0.110} \\\hline
    \end{tabular}
    \label{table:self}
\end{table}

\subsection{Ablation Study} \label{sec:ablation}
Table~\ref{table:ablation} provides a detailed ablation study on KITTI Stereo 2015 dataset. "w/o expansion" denotes the result of feature enhancement only on the locations with sparse cues. Contrarily, "w/ expansion" stands for applying feature enhancement on all the expanded area with same Gaussian function (with no distance weighted). Comparing the two rows verifies the effectiveness of sparsity expansion. Even though without distance weighted feature enhancement, enlarging the guided area by sparsity expansion brings significant gain. Adding distance weighted function or the shifted version increase even more performance. For the comparison of the two distance weighted function $f$ and $f^s$, $f^s$ is better than $f$ in the most of the case.

\begin{table}[]
    \centering
    \caption{Ablation study on KITTI 2015: We can observe the huge performance boost with expansion in both backbones and fine tune scenarios. Compared with naive expansion, distance weighted feature enhancement can further improve the performance.}
    \begin{tabular}{lcc|cc}
         \multirow{2}{*}{Model} & \multicolumn{2}{c|}{w/o Fine Tune} & \multicolumn{2}{c}{Fine Tune} \\ \cline{2-3} \cline{4-5} 
         & PSMNet & GANet & PSMNet & GANet\\ \hline \hline
         Baseline&8.568&3.238&0.764&0.743\\
         w/o expansion&7.544&2.956&0.603&0.586\\
         w/ expansion&5.189&2.751&0.506&\textbf{0.506}\\
         w/ distance weighted $f$&5.122&2.619&0.542&\textbf{0.506}\\
         w/ distance weighted $f^s$&\textbf{4.624}&\textbf{2.030}&\textbf{0.502}&0.512\\
    \end{tabular}
    \label{table:ablation}
\end{table}
\subsection{Robustness over Data Sparsity} \label{sec:sparsity test}
\begin{figure}
    \centering
    \includegraphics[width=0.5\textwidth]{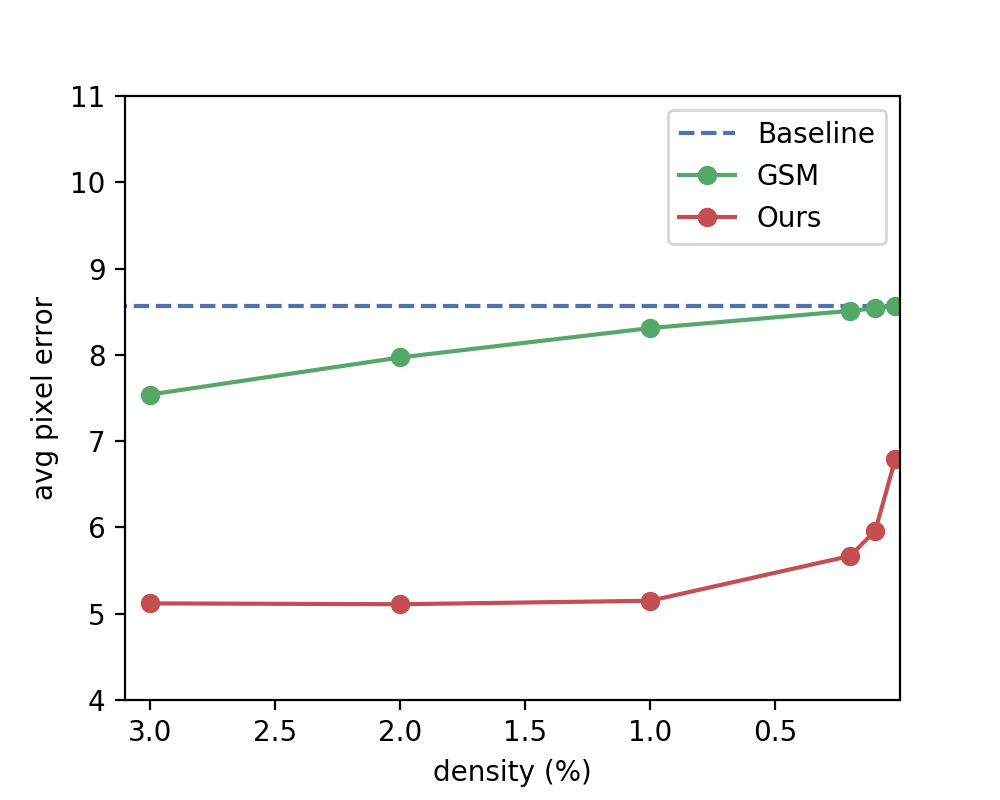}
    \caption{Robustness over sparsity. We progressively adjust the sampling density of the LiDAR points on KITTI Stereo 2015. The average pixel error increases as the density of LiDAR points drops. This shows our method has better robustness when the points are extremely sparse. See more in section \ref{sec:sparsity test}.}
    \label{fig:density}
\end{figure}

We progressively adjust the sampling density of the LiDAR point cloud to generate sparser guidance. We can see the result in Figure~\ref{fig:density}. At the beginning, when density is $3\%$, both GSM \cite{poggi2019guided} and our method are better than the baseline model. However, as the density gets lower, the error of GSM become very close to the baseline model since there are no enough locations for GSM to apply feature enhancement. In contrast, our method hardly drop until the density is extremely sparse, smaller than $1\%$. This emphasizes the robustness of our methods.

\subsection{Visualization Results}
\begin{figure}
    \centering
    \includegraphics[width=\textwidth]{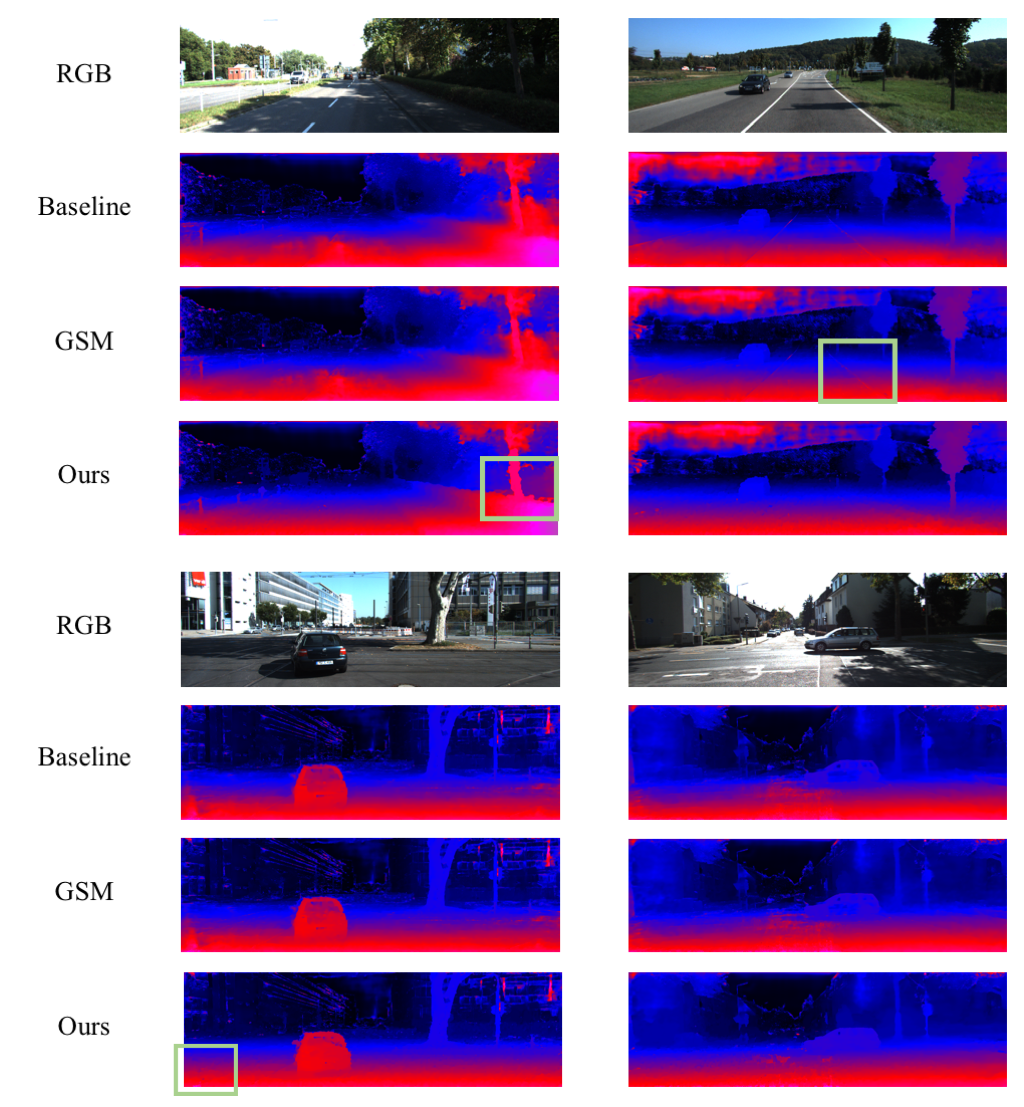}
    \caption{Visualization results on KITTI 2015 stereo (with pretrain) with GANet as backbone. Leveraging sparse but accurate LiDAR cues can boost the performance of stereo matching in terms of the difficult lighting condition (bottom right) and repetitive patterns (bottom left). With sparsity expansion, the performance can be extraordinarily boosted via similar scene structure (top left).}
    \label{fig:vis-ga}
\end{figure}

In Figure~\ref{fig:vis-ga}, it shows the visualization results compared to other methods with GANet as backbone. In the top left result, the sparse cues can expand and enhance disparity estimation even when RGB image is dark and blurred. In the top right example, the pretrained network is mislead by the white line on the road surface. Our feature enhancement methods can adjust the errors by leveraging expansion of sparse LiDAR with respect to color intensity. The bottom left indicates the smoothness ability of the feature enhancement. 

\begin{figure}
    \centering
    \includegraphics[width=\textwidth]{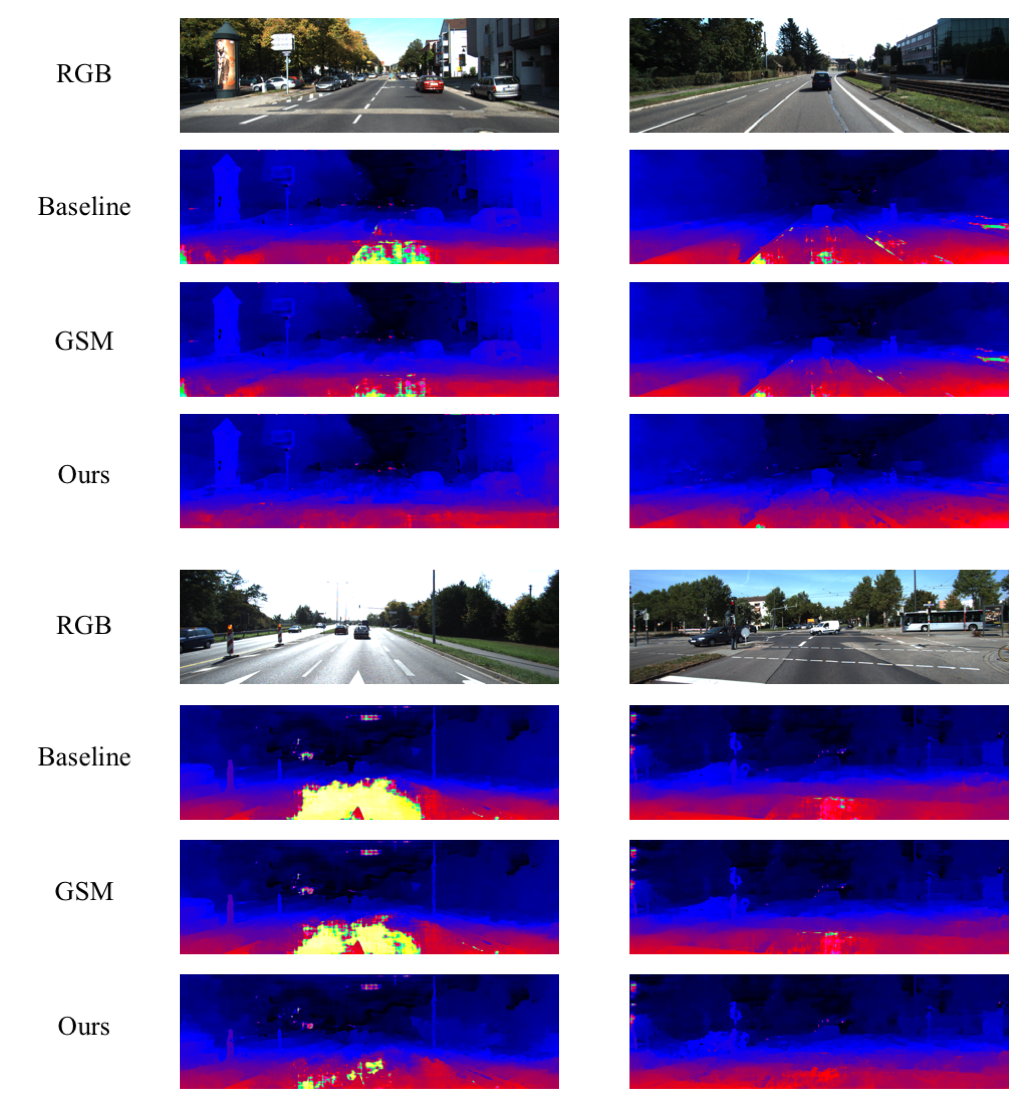}
    \caption{Visualization results on KITTI 2015 stereo (with pretrain) with PSMNet as backbone. The results show that with feature enhancement from expanded disparity can largely compensate for the error of stereo matching even under huge lighting variance or repetitive patterns.}
    \label{fig:vis-psm}
\end{figure}

In Figure~\ref{fig:vis-psm}, the bottom left example demonstrates the situation when stereo matching algorithm suffers from environment lighting. Our feature enhancement method overcomes the weakness of stereo matching, and prove the idea that we can benefit from the complementary characteristics of stereo matching and sparse LiDAR is true. 

\section{Conclusions}
We propose sparsity expansion, a new technique 
to leverage sparse cues as the guidance for stereo matching algorithms in this work. We search for the effective area of each guidance points instead of only cares about the location with sparse cues. This approach shows robustness when the data is highly sparse. We also propose distance weighted feature enhancement that regulates the level of enhancement according to its distance to the guidance point. Taking sparse LiDAR points as the guidance of deep stereo matching, the networks is able to reduce the domain shift issue with or without additional fine-tuning. 
To show the effectiveness of our proposed method, we do quantitative and qualitative experiments to demonstrate the idea on several datasets and multiple deep stereo networks. The performance boost are huge with respect to feature enhancement. We also show the success of the method for improving the model trained by self-supervised learning.

On the whole, we take the effective areas of sparse guidance into consideration to formulate the sparse guided stereo matching problem. In addition, we propose sparsity expansion and distance weighting to ameliorate the performance. Finally, our idea is deployable to existing stereo matching algorithms with cost volume.

In the future, we seek to open up the possibility for other task that also suffered from data sparsity issue.

\section{Acknowledgement}
This work was supported in part by the Ministry of Science and Technology, Taiwan, under Grant MOST 109-2634-F-002-032 and FIH Mobile Limited. We are grateful to the National Center for High-performance Computing.



\clearpage
%
%

\clearpage

\title{Expanding Sparse Guidance for Stereo Matching (Supplementary Material)}
\maketitle

We will provide more experiment results and comparison for further discussion in supplementary material.
\section{Evaluation on Middlebury v3}
\begin{figure}
    \centering
    \includegraphics[width=\textwidth]{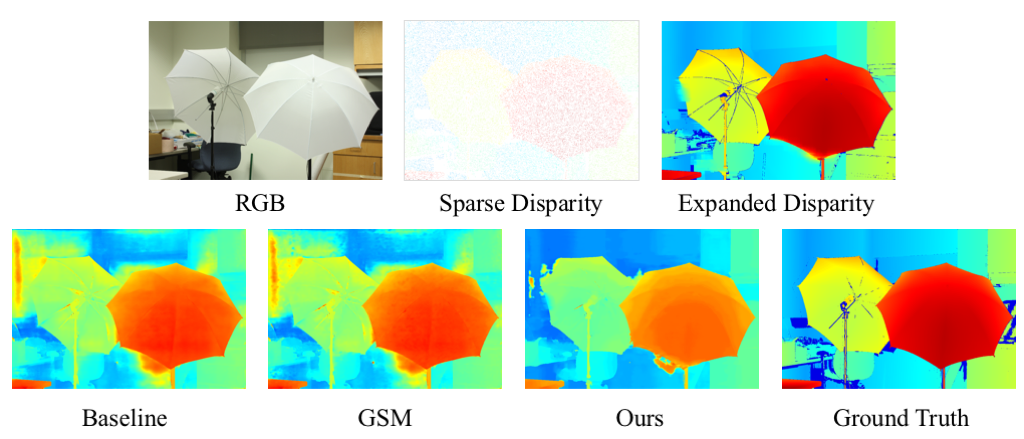}
    \includegraphics[width=\textwidth]{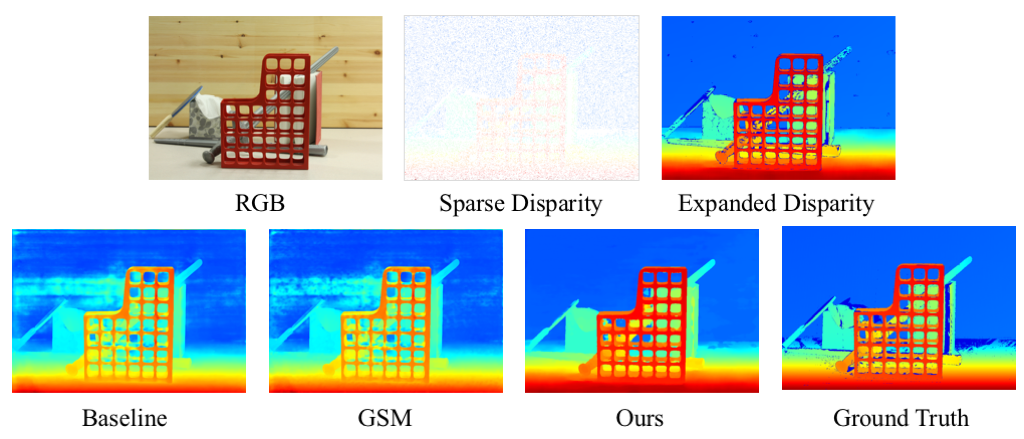}
    \includegraphics[width=\textwidth]{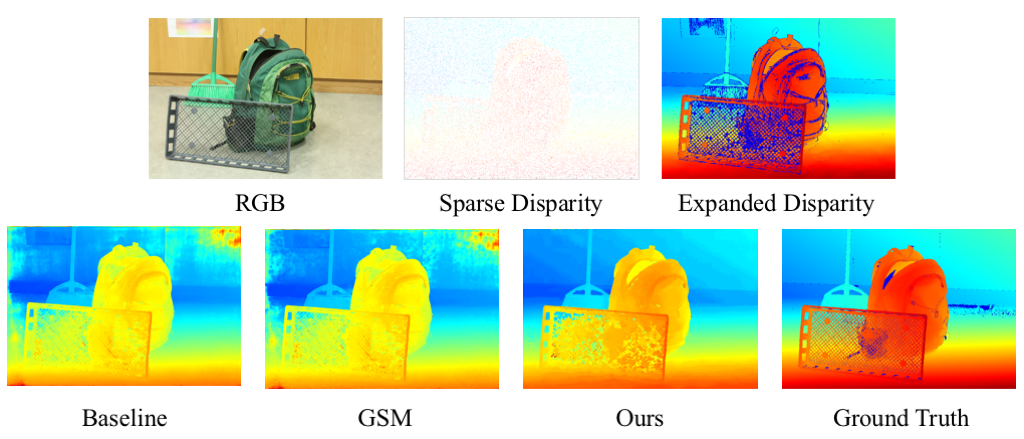}
    \caption{Middlebury v3 visualization results without fine-tuning. It is worth to notice that the expanded disparity is dense and similar to ground truth. With the expanded disparity, the output disparity maps are enhanced especially in the background.}
    \label{fig:middlebury}
\end{figure}

We also evaluate our methods on Middlebury v3~\cite{scharstein2014high}. Middlebury v3 is composed of 23 stereo pairs with dense ground truth disparity. We subsample $5\%$ disparity value from ground truth disparity map as the sparse disparity map. We test on Middlebury with   PSMNet \cite{chang2018pyramid} and GANet \cite{zhang2019ga} pretrained on SceneFlow \cite{mayer2016large} without fine-tuning. The results are shown in Table \ref{table:middlebury}. The visualization results are shown in Figure \ref{fig:middlebury}.

\setlength{\tabcolsep}{2mm}{
\begin{table}
    \centering
    \caption{Evaluation results on Middlebury v3.  "Baseline" means model inferences directly with pretrained SceneFlow \cite{mayer2016large} weights. Our method outperforms previous methods.}
    \begin{tabular}{|c|l||c|cccc|}\hline
         \multirow{2}{*}{Backbone}&\multirow{2}{*}{Model}&\multirow{2}{*}{Avg PX}&\multicolumn{4}{c|}{Error Rate}  \\ \cline{4-7}
         &&&$>2$&$>3$&$>4$&$>5$\\ \hline \hline 
         \multirow{3}{*}{PSMNet\cite{chang2018pyramid}}&Baseline&9.594 & 0.803 & 0.697 & 0.606 & 0.530 \\
         &GSM\cite{poggi2019guided}&9.553 & 0.796 & 0.693 & 0.606 & 0.532\\
         &\textbf{Ours}& \textbf{6.924} & \textbf{0.738} & \textbf{0.634} & \textbf{0.540} & \textbf{0.457} \\\hline \hline
         
         \multirow{3}{*}{GANet\cite{zhang2019ga}}&Baseline& 5.859 & 0.484 & 0.316 & 0.244 & 0.205\\
         &GSM\cite{poggi2019guided}&5.570 & 0.436 & 0.277 & 0.216 & 0.184\\
         &\textbf{Ours}& \textbf{3.567} & \textbf{0.243} & \textbf{0.100} & \textbf{0.072} & \textbf{0.061} \\\hline
    \end{tabular}
    \label{table:middlebury}
\end{table}
}

\section{Ablation Study of Sparsity Expansion}
In previous ablation study section, we have proved that applying expansion and different distance weightings are effective. This section will focus on different expanding approaches and explain our choice for expansion technique.
\subsection{Expansion Region}
One naive baseline is to expand sparsity with square block. To be specific, we expand the sparse guidance to a $L' \times L'$ square area with the center as the guidance point. Overlapped regions are filled with interpolated value according to distance from the guidance point. To be fair, we set $L = L'$ to compare square expansion with our expansion approach in the experiments. In Table \ref{table:ablation-square}, the performance of our expansion technique is better.

\begin{table}
    \centering
    \caption{Comparison of expansion region type. We compare our method with naive square expansion. The performance demonstrates our sparsity expansion based on RGB hints is reasonable.}
    \begin{tabular}{lc|cc}
         Model&Region Type&KITTI 2012&KITTI 2015  \\ \hline \hline
         w/ expansion & Square & 5.884 & 5.399 \\
         \textbf{w/ expansion} & \textbf{Ours} & \textbf{5.883} & \textbf{5.189} \\ \hline
         w/ distance weighted $f$& Square & 5.898 & 5.399 \\
         \textbf{w/ distance weighted $f$}&\textbf{Ours}&\textbf{5.866}&\textbf{5.122}\\ \hline
         w/ distance weighted $f^s$& Square & 5.512 & 5.182 \\
         \textbf{w/ distance weighted $f^s$}&\textbf{Ours}&\textbf{5.273}&\textbf{4.624}\\
    \end{tabular}
    \label{table:ablation-square}
\end{table}

\subsection{Segmentation Based Expansion} \label{sec:segmentation}
Expanding sparsity maps with respect to semantics may sound more reasonable and meaningful. However, we argue that the trade of between performance and time complexity is important. In Table \ref{table:ablation-segmentation}, we compare our methods with segmentation-based method SLIC \cite{achanta2012slic}. 

SLIC \cite{achanta2012slic} uses k-means to find the segment regions in an iterative way. If there are more than one guided points in the same segmentation region, we fill up the region by the averaged disparity value in the region. The time complexity depends on the number of initial seeds $k$, searched square distance $S$ and iterative time $I$. Specifically, the total time complexity is $O(k S^2 I)$. 

Compared to our techniques, our worst case of time complexity is  $O(NL^2)$, where $N$ is the number of guidance and $L$ is the expansion limits. The time complexity is smaller than SLIC by a factor of iterative times $I$ when the number of initial seeds are same with the guidance points ($k=N$) and the searched distance is the same as the maximum expansion distance $S=L$. 

We show why our expansion technique is preferred compared with segmentation-based methods. In Table \ref{table:ablation-segmentation}, we compare our expansion method with segmentation-based SLIC \cite{achanta2012slic}. We set number of segments to be $3200$ to densely cover the sparse disparity map. SLIC slightly surpasses our pure expansion techniques, which fulfills our expectation that segmentation-based methods can be more precise for expansion. However, when time complexity is taken into consideration, our methods are faster and more suitable for real-time autonomous driving usage. In practice, our techniques can be highly paralleled because the expansion centers are independent of each other, while iterative methods must be done in sequence. Furthermore, our expansion technique outperforms SLIC when leveraging distance weighted enhancement. 


\setlength{\tabcolsep}{1.2mm}{
\begin{table}
    \centering
    \caption{Comparison of segmentation-based expansion and ours on KITTI 2012. "Time Complexity" is the pre-processing time complexity for disparity expansion. The results show the trade-off between performance and speed. With our method, applying distance weighted function can bridge the performance gap and even surpass the segmentation-based method both in time and performance. In real-time environment (e.g., autonomous driving), our technique is more appealing than segmentation-based one. Details are presented in Section~\ref{sec:segmentation}.}
    \begin{tabular}{l|c|c|cccc}
         \multirow{2}{*}{Model}&Time&\multirow{2}{*}{Avg PX}&\multicolumn{4}{c}{Error Rate}  \\ \cline{4-7}
         & Complexity & & $>2$ & $>3$ & $>4$ & $>5$ \\ \hline \hline
         w/o expansion & - & 7.319 & 0.784 & 0.678 & 0.577 & 0.485\\ 
         seg-based expansion & $O(kS^2I)$ & 5.733 & 0.781 & 0.682 & 0.589 & 0.498\\ 
         our expansion & $O(NL^2)$ & 5.883 & 0.781 & 0.683 & 0.592 & 0.505\\
         \textbf{our distance weighted} & $O(NL^2)$ & \textbf{5.273} & \textbf{0.753} & \textbf{0.645} & \textbf{0.544} & \textbf{0.449}\\ 
    \end{tabular}
    \label{table:ablation-segmentation}
\end{table}
}

\section{Comparison with Pseudo LiDAR++}
Pseudo LiDAR++ \cite{you2019pseudo} proposes graph-based depth correlation (GDC) algorithm in order to improve the result of 3D object detection. 
Given the sparse ground truth 3D point cloud data (e.g., LiDAR), by projecting the output depth map into 3D point cloud (pseudo LiDAR), a neighborhood relation graph can be formed via KNN. Then, depth values can be corrected by a graph-based optimization process. 

Although the neighbor regions of the guidance are taken into account, exploring the neighborhood relation in 3D space can be slow. The optimization could become a bottleneck with denser points. In contrast, we directly apply feature enhancement by utilizing the physical meaning of cost volume. Our method is well suitable for end-to-end fine-tuning and can be highly parallelized.

We take the official code of GDC released by \cite{you2019pseudo} and evaluate on KITTI Stereo 2015 dataset. We apply GDC on the output of PSMNet trained with SceneFlow and PSMNet with fine-tuning. The output disparity map is first converted into depth map and then projected into 3D space with calibration parameters provided by the dataset. The sparse ground truth data (which is also disparity map) is projected into 3D by the same approach. We then turn the corrected depth values back into disparity values for evaluation.

\begin{table}[h]
    \centering
    \caption{Comparison with GDC in pseudo LiDAR++ \cite{you2019pseudo} on KITTI stereo 2015 and PSMNet. Given extra sparse ground truth data, our method is able to better utilize this information and achieve better performance with or without fine-tuning.}
    \begin{tabular}{|c|c|c|cccc|} \hline
         \multirow{2}{*}{Model}&\multirow{2}{*}{Fine Tune}&\multirow{2}{*}{Avg PX}&\multicolumn{4}{c|}{Error Rate}\\ \cline{4-7}
         & & & $>2$ & $>3$ & $>4$ & $>5$ \\ \hline \hline
         Baseline & & 8.568 & 0.730 & 0.600 & 0.486 & 0.390 \\
         Pseudo Lidar ++ & & 7.132 & \textbf{0.592} & \textbf{0.490} & \textbf{0.399} & \textbf{0.321}\\
         \textbf{Ours} & & \textbf{4.624} & 0.713 & 0.584 & 0.472 & 0.373\\ \hline
         Baseline & \checkmark & 0.764 & 0.055 & 0.029 & 0.019 & 0.014\\
         Pseudo Lidar ++ & \checkmark & 0.597 & 0.043 & 0.022 & 0.015 & 0.011\\
         \textbf{Ours} & \checkmark & \textbf{0.502} & \textbf{0.023} & \textbf{0.012} & \textbf{0.008} & \textbf{0.006}\\ \hline
    \end{tabular}
    \label{table:pseudo-lidar++}
\end{table}

The result is shown in Table~\ref{table:pseudo-lidar++}. Without additional fine-tuning, our method outperforms GDC in average pixel error. Compared to GDC, our method is able to take larger neighbor areas into consider because of the simplicity design of sparse guidance expansion. GDC handles the structure detail better due to the correction in 3D space and careful optimization, resulting lower $n$-pixel error. With fine-tuning PSMNet on KITTI dataset, apply GDC on the output of network is able to lower the error. On the other hand, our method can be end-to-end trained in the fine-tune stage and further improve the result, including average pixel error and $n$-pixel error.

\section{More Visualization Results}
\begin{figure}
    \centering
    \includegraphics[width=0.8\textwidth]{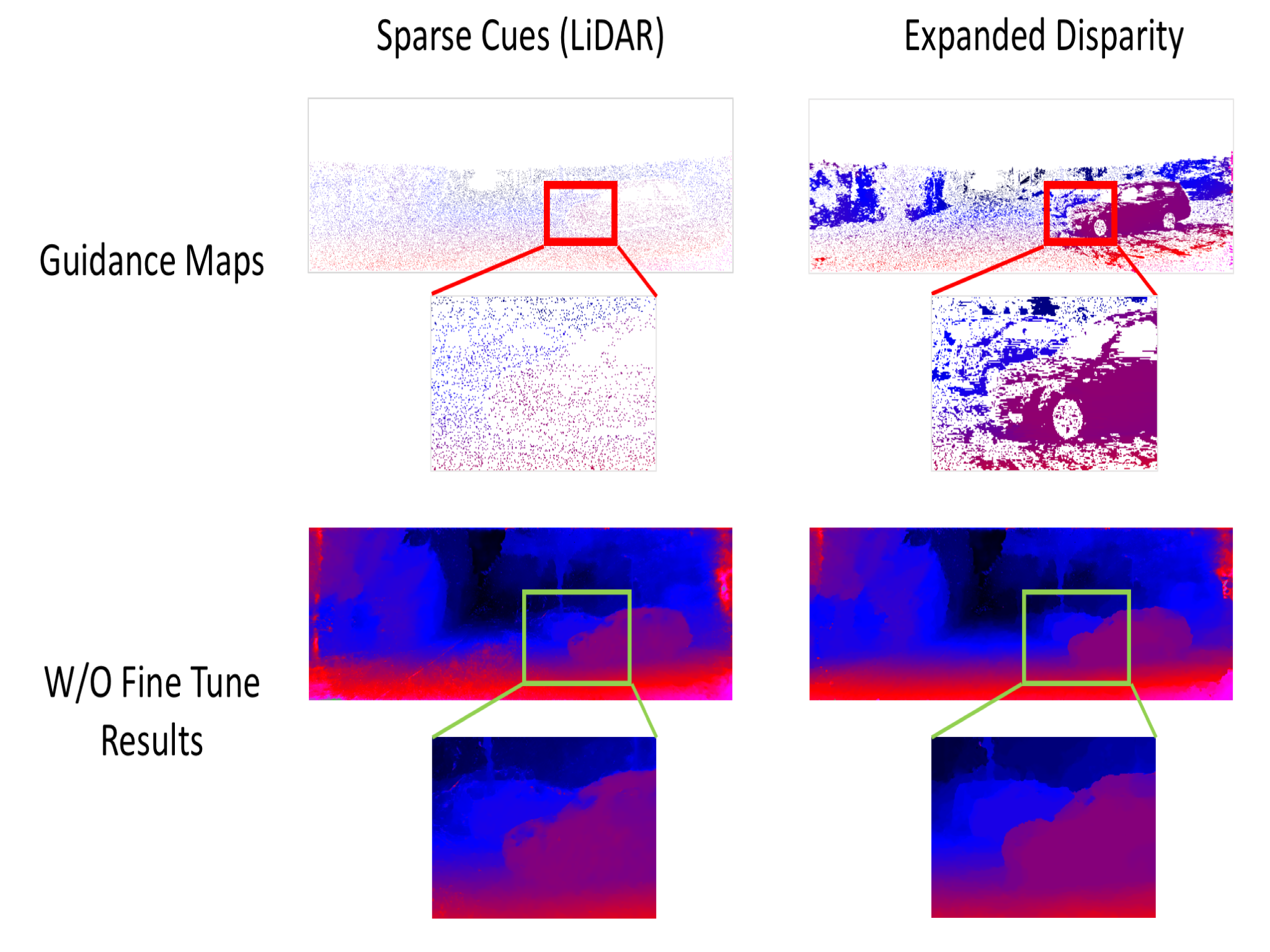}
    \caption{Visualization on the detailed parts. The improvement comes from the expanded disparity.}
    \label{fig:more-vis}
\end{figure}

\clearpage
%
%
\bibliographystyle{splncs04}
\bibliography{egbib}
\end{document}